\documentclass[11pt]{article}

\usepackage{acl}
\usepackage{times}
\usepackage{latexsym}
\usepackage{authblk}
\usepackage{booktabs}
\usepackage{hhline}
\usepackage{longtable}
\usepackage[T1]{fontenc}

\usepackage[utf8]{inputenc}

\usepackage{microtype}

\usepackage{inconsolata}

\usepackage{graphicx}

\title{Can Finetuing LLMs on Small Human Samples Increase Heterogeneity, Alignment, and Belief-Action Coherence?}

\author[1]{Steven Wang\thanks{Corresponding author: swang57@buffalo.edu}}
\author[1]{Kyle Hunt}
\author[1]{Shaojie Tang}
\author[2]{Kenneth Joseph}
\affil[1]{Department of Management Science and Systems, University at Buffalo, Buffalo, NY, USA}
\affil[2]{Department of Computer Science and Engineering, University at Buffalo, Buffalo, NY, USA}

\date{\today}

\begin{document}
\maketitle
\begin{abstract}
There is ongoing debate about whether large language models (LLMs) can serve as substitutes for human participants in survey and experimental research. While recent work in fields such as marketing and psychology has explored the potential of LLM-based simulation, a growing body of evidence cautions against this practice: LLMs often fail to align with real human behavior, exhibiting limited diversity, systematic misalignment for minority subgroups, insufficient within-group variance, and discrepancies between stated beliefs and actions. This study examines an important and distinct question in this domain: whether finetuning on a small subset of human survey data, such as that obtainable from a pilot study, can mitigate these issues and yield realistic simulated outcomes. Using a behavioral experiment on information disclosure, we compare human and LLM-generated responses across multiple dimensions, including distributional divergence, subgroup alignment, belief–action coherence, and the recovery of regression coefficients. We find that finetuning on small human samples substantially improves heterogeneity, alignment, and belief–action coherence relative to the base model. However, even the best-performing finetuned models fail to reproduce the regression coefficients of the original study, suggesting that LLM-generated data remain unsuitable for replacing human participants in formal inferential analyses. 
\end{abstract}

\section{Introduction}
With the rapid advancement of reasoning and language capabilities in large language models (LLMs), researchers have increasingly explored their use as human surrogates to simulate survey and experimental responses. One of the earliest works in this direction is by \citet{argyle2023out}, who used GPT-3 to model public opinion patterns in political science. Since then, the use of LLMs as proxies for human participants has expanded across a range of disciplines, including marketing, economics, and psychology \cite[e.g.][]{brand2023using, aher2023using}, as well as in natural language processing (NLP) tasks traditionally performed by humans, such as text quality evaluation and data annotation \citep{Chiang2023CanLL,he2024annollm}. In parallel, another line of work has focused on finetuning open-source LLMs with human experimental data to better align model predictions with real behavioral distributions. For example, Centaur \citep{binz2025foundation} is an LLM finetuned on a large-scale psychology dataset to more accurately mimic  differences in human behavior, while SubPOP \citep{suh2025language} leverages real survey data to model subpopulation opinion distributions. Together, these developments have spurred growing interest in whether LLMs can serve not only as reasoning systems but also as credible simulators of human heterogeneity and social behavior. 

On the other hand, a growing body of research urges caution in using LLMs as substitutes for human participants. The central critique is that LLM-generated responses lack heterogeneity, tending to reproduce human averages while underrepresenting the diversity found in real populations \citep{bisbee2024synthetic, wang2025large, wu2025llm}. This issue is most pronounced under default prompting conditions, where LLMs do not behave like an “average” respondent but instead reflect the perspectives of the WEIRD demographic (Western, Educated, Industrialized, Rich, and Democratic) that dominates their training data \citep{liu2022quantifying}. To address this limitation, researchers have experimented with persona and context-based prompting, embedding demographic cues such as gender, race, or age to encourage more varied responses. While these methods introduce some between-group differences, they remain imperfect. Studies have also documented the phenomenon of identity flattening, where LLMs exhibit variation across demographic subgroups but not within them, portraying each subgroup as internally homogeneous \citep{lin2025six, wang2025large}. This behavior often mirrors out-group stereotyping, where the model reproduces what outsiders might expect a given group to believe, rather than authentic in-group diversity, reflecting biases inherent in online text corpora. Even finetuned models designed specifically for psychological realism face similar challenges. For example, Centaur \citep{binz2025foundation} has been found to deviate systematically from human ratings of moral judgment \citep{schroder2025large} and behavioral study data \citep{Gao}, suggesting limited generalizability to out-of-distribution behavioral contexts.

The aim of this study is not to enter the discourse of whether off-the-shelf or pre-finetuned LLMs can fully substitute for human respondents, but rather to examine whether finetuning on a small subset of human data, such as that obtainable from a pilot study, can mitigate the limitations documented in prior work. For reasons detailed below, we base our simulations on the behavioral experiment of \citet{Hunt}, which investigates how information disclosure shapes attackers’ judgments and decisions in a security scenario. Our findings show that LLMs finetuned on small fractions of the human dataset can substantially improve heterogeneity, distributional alignment, and value–action coherence relative to the base LLM. However, despite these improvements, none of the finetuned models reliably reproduce the original regression coefficients or underlying hypothesis tests, indicating that replacing human participants with LLMs remains risky and may introduce non-negligible bias into conclusions. Hence, we caution against the naive use of base models for experimental design and simulation, even when augmented with sociodemographic prompting, they tend to produce overly tight variance. Such compression can yield overconfident power analyses and lead to underpowered studies. In contrast, we show that finetuning on as few as 30 human observations substantially improves distributional alignment, producing variation that more closely resembles human data and yielding more realistic simulated responses. This makes finetuned models better suited for certain stages of survey design and pretesting than base LLMs. In sum, we contribute to the literature by showing that finetuning on a small sample of human data (30 observations) can substantially mitigate several issues identified in prior work, including the lack of heterogeneity, distributional misalignment, and the value–action gap. Both the original and simulated data will be released for future research upon acceptance. 

\section{Related Literature}
\subsection{Limitations of LLMs in Replicating Human Behavior: Homogeneity, Misalignment, and the Value-action Gap}
A growing body of research shows that LLMs often fail to capture the full heterogeneity of human responses when used to simulate survey participants \citep{wang2025large, hewitt2024predicting, dominguez2024questioning, wu2025llm, bisbee2024synthetic, giorgi2024modeling}. \citet{bisbee2024synthetic} examine this limitation within political science using simulations based on the American National Election Study. They find that while LLMs can accurately reproduce mean responses, they are systematically biased, overconfident, and display artificially low variance. \citet{wang2025large} extend this line of research across multiple policy domains, such as health care, gun regulation, immigration, and criminal justice, using prompts that represent sixteen demographic identities. Their results reveal a pronounced lack of within-group variance: responses generated for each demographic subgroup are strikingly homogeneous. The authors describe this as identity flattening, whereby the model constructs one-dimensional, stereotyped “personalities” for each demographic. \citet{Gao} examine the alignment between human participants and LLM-simulated responses through controlled experiments. They report that this lack of heterogeneity and behavioral misalignment persists even after applying various mitigation strategies, including few-shot prompting, role or persona assignment, multilingual prompting, and retrieval-augmented generation (RAG) \citep{lewis2020retrieval}. \citet{shen2025mind} draw on the concept of the value–action gap from the social science literature \citep{blake1999overcoming, godin2005bridging, chung2007value}, which refers to the discrepancy between individuals’ stated values and their actual behaviors. They apply this framework to examine the gap between an LLM’s expressed inclinations and its subsequent decisions. Collectively, these studies demonstrate that while LLMs can replicate average human judgments with reasonable accuracy, they systematically underrepresent the variance and contextual diversity that characterize real human populations. 

\subsection{Attempts to Enhance the Behavioral Realism of LLMs}
 Aiming to generate more diverse and contextually grounded outputs, researchers have increasingly incorporated persona or role-based prompting when simulating human responses \citep{argyle2023out, dillion2023can}. On standard reasoning and coding benchmarks, persona prompts (e.g., “act as a math teacher/engineer”) and more advanced steering frameworks have shown modest but consistent improvements in performance, suggesting that explicit role framing can increase reasoning and improve alignment with task goals \citep{wang2025improving, kim2024persona, kong2024better}. However, when applied to survey or opinion-elicitation settings, often referred to as sociodemographic prompting, these techniques produce mixed or negative results. Studies find that large language models tend to misrepresent marginalized groups and align more closely with white or majority identities than with minority participants, generating stereotyped or homogenized responses that fail to capture genuine within-group heterogeneity \citep{lutz2025prompt, sun2025sociodemographic, wang2025large}.

\subsection{Gaps in Existing Work}
Despite substantial work documenting the limitations of LLMs as simulated survey respondents, very little research has explored whether these issues can be mitigated through targeted finetuning on small samples of human data. Existing studies typically evaluate base models or rely on prompt engineering, or persona conditioning, but do not examine whether a small, realistic amount of human data such as that obtainable from a pilot study can meaningfully improve heterogeneity, alignment, and behavioral coherence. This gap leaves open the critical question of whether modest finetuning can address the flaws identified in earlier research, or whether these limitations are inherent to current LLM architectures.

\section{Data}
This study builds on a large behavioral experiment by \citet{Hunt}, which examined the effects of information disclosure on attackers' judgments and decisions in a security setting. Specifically, the experiments studied how information disclosed about the deployment of new security technology impacts attackers' beliefs regarding \textit{where} the technology is deployed and their final attack decisions (i.e., whether to attack, and if so, where). We selected this experiment for use in the current research for three main reasons. First, the design of the experiment, which elicited both beliefs and decisions from the participants, enables us to assess the coherence between LLMs’ stated beliefs and their subsequent actions. Second, we obtained the original data directly from the authors of the experiment, with confirmation that the data had not yet been publicly released or fed into any LLM prior to the analysis presented herein. Ensuring that the model had not previously seen the dataset is critical, as prior work demonstrates that LLMs can memorize training data, including personally identifiable information, and reproduce it verbatim with only one related document in the training corpus \citep{zhou2024quantifying, carlini2021extracting, carlini2022quantifying}. Such data contamination undermines the validity of LLM evaluations by producing spuriously high or unreliable performance estimates \citep{sainz2023nlp, li2024open, ranaldi2024investigating}. Third, the experiments in \citet{Hunt} focused on information disclosure in a high-stakes, strategic environment (i.e., security and defense), a domain that has not been studied in the context of LLMs and behavioral realism.

The sample we analyze consists of behavioral data from 929 U.S. adults \footnote{The original dataset in \citet{Hunt} had 975 responses, but we exclude data points with missing demographic information for both finetuning and inference.} who played the role of attackers deciding whether to attack either of two targets that were defended by a centralized defender. In the experiment, the defender released one of four disclosures related to the deployment of new security measures to protect her assets at the targets. 
Participants (i.e., attackers) were also informed of the defender's relative valuations of both targets, where the differences between her valuations of each target could be large or small. Appendix \ref{sec:manipulations} outlines the manipulations in more detail. This 2 x 4 design created a total of eight treatment groups (i.e., two levels of the valuation factor, and four levels of the disclosure factor). After receiving the experimental manipulations, participants then reported their beliefs regarding where the technology was deployed -- expressed as probabilities that the new security measures were deployed at Target 1, Target 2, or both, and then made a final decision on whether to attack Target 1, attack Target 2, or not attack. Participants were incentivized with base pay for completing the study and potential bonus payments based on the outcomes of their attacks. The full experimental script can be found in Appendix \ref{sec:prompt}.

\section{Methods}
Our goal is to explore how well various LLM-based approaches can predict various results derived from this experimental dataset, and in particular how an approach finetuned on a subset of this data (e.g. that would come from a pilot study) can improve these predictions. We consider three different approaches, described below. All simulations were conducted using GPT-4.1 \citep{achiam2023gpt} with the default temperature setting of 1.0. The finetuned models use the same base architecture and temperature configuration to ensure comparability. 
We mimic most prior work discussed above and first ran simulations using an unfinetuned GPT model. We incorporated participant demographics as role prompts to approximate more realistic responses. For finetuning, we used two sampling strategies and three dataset sizes, resulting in six models in total. The first strategy uses random sampling, and the second strategy, which we refer to as balanced sampling, draws equal proportions \footnote{We use a round-robin sampling strategy in which we randomly draw one sample from each ethnicity group in sequence until reaching 25\% of the dataset. For ethnicity categories with smaller sample sizes (e.g., Other), we include all available observations and then continue the round-robin procedure with the remaining groups to reach the target proportion.} of responses from each ethnic subgroup in order to address the systematic misalignment for minority groups noted in prior research.  The three dataset sizes correspond to the number of treatment groups included in finetuning. This design allows us to test whether finetuning on only a subset of treatment groups enables the model to generalize to the remaining groups. Specifically, we sample 25\% of the data from the first treatment group (1a1b), from the first four treatment groups, and from all eight treatment groups. For the multi group settings, we draw 25\% from each included group. We name each model using its sampling strategy and dataset size. For example, the model finetuned on 25\% of the first treatment group using random sampling is labeled Random (1a1b), with parallel naming for the remaining random and balanced models.

As outcomes, we examine how finetuning affects heterogeneity with just 25 \% of the data from a single treatment group (the 1a1b treatment group; see Appendix \ref{sec:manipulations}). To quantify distributional alignment, we use the Jensen–Shannon (JS) distance \citep{lin2002divergence}, a symmetric and bounded measure of distributional divergence that has also been applied by \citet{Gao} to compare LLM and human response distributions. For belief–action gap, we measure the conditional proportion of LLMs choosing to attack the target that they think has the highest probability that security is in place and compare it to humans. Lastly, we compute the accuracy of regression coefficient estimates relative to the regression analysis performed on the human data.  

\section{Results}
In the following subsections, we examine how finetuning affects heterogeneity, distributional alignment, and the belief–action gap, as well as the accuracy of regression coefficient estimates relative to the regression analysis performed on the human data.
\subsection{Heterogeneity}

\begin{figure*}
    \centering
    \includegraphics[width=0.48\textwidth]{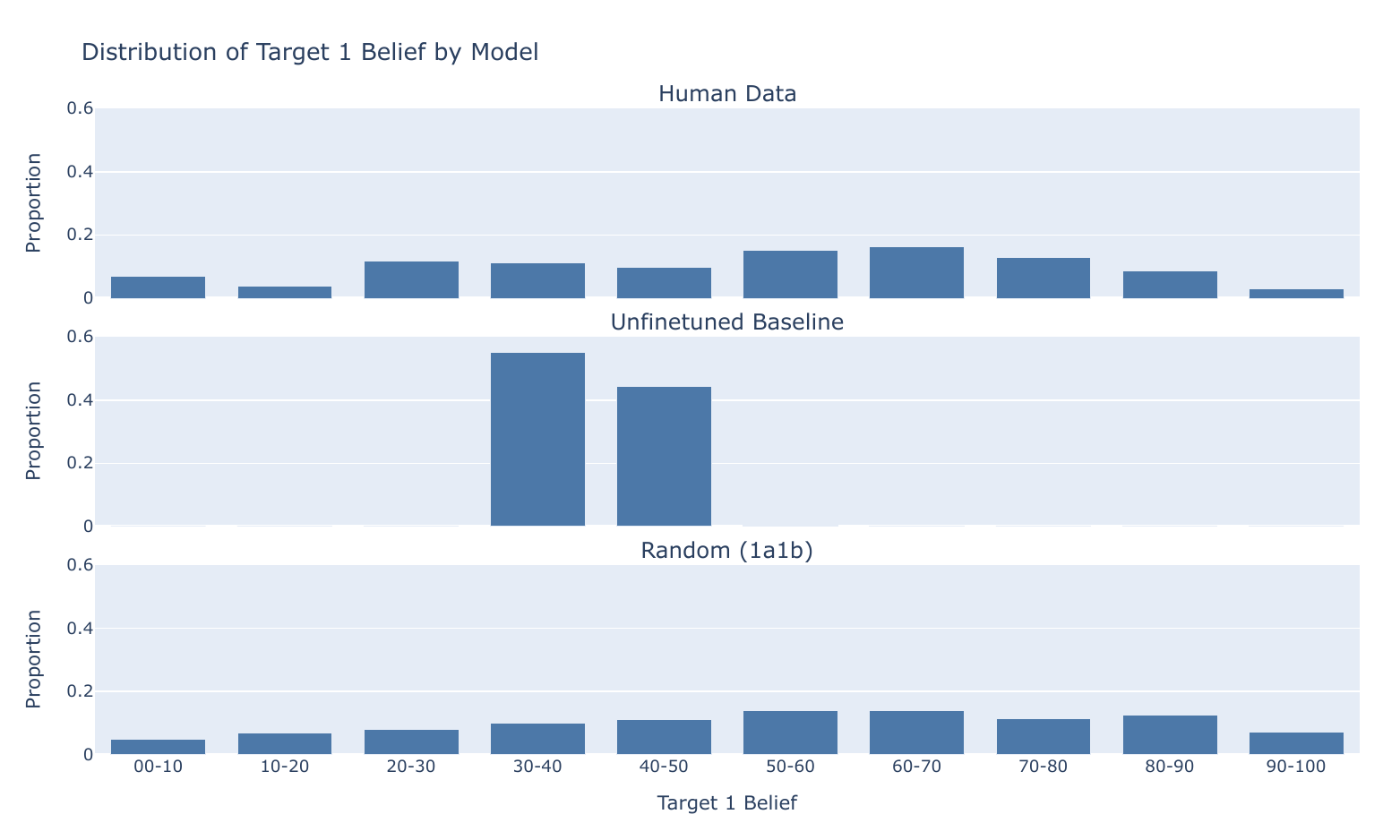}\hfill
    \includegraphics[width=0.48\textwidth]{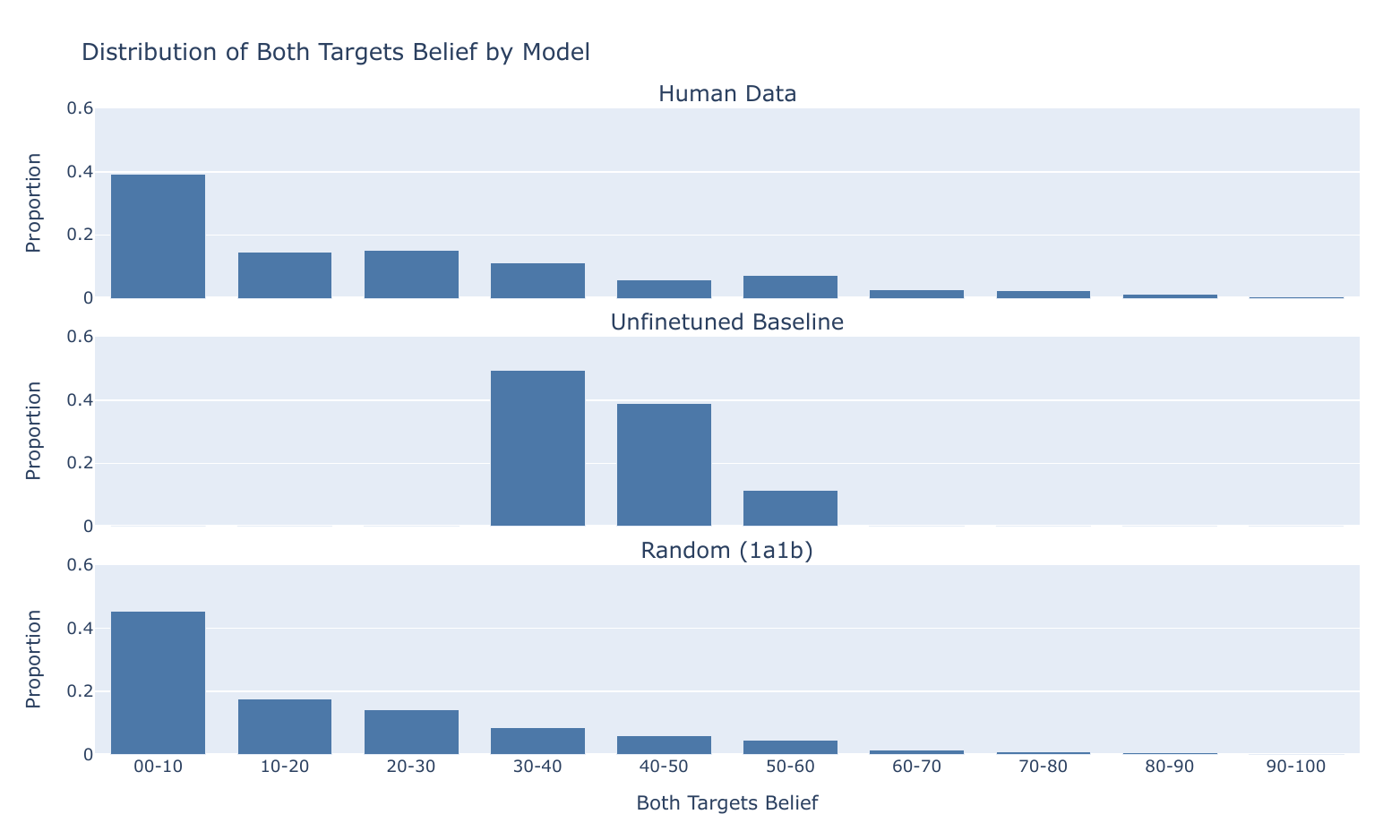}
    \caption{Distributions of beliefs that security is deployed at Target 1 (left) and at both targets (right) across human data and LLM simulation conditions. Each panel shows the empirical belief distribution for human participants alongside models finetuned under different sampling strategies.}
    \label{fig:b1_b3_dist}
\end{figure*}

We find that even a small number of finetuned samples can drastically improve diversity and mitigate homogeneity issues relative to the unfinetuned baseline. Figure \ref{fig:b1_b3_dist} shows the distribution of beliefs regarding the likelihood that the defender’s security measures were deployed at Target 1 and both Targets. The analysis of Target 2’s beliefs and the results from all models are presented in Appendix \ref{sec:appendix_belief}, and support findings discussed here. In analyzing Target 1 (left panel) in Figure \ref{fig:b1_b3_dist}, the unfinetuned baseline exhibits a highly concentrated distribution, with most responses clustered in the 30–40 \% and 40–50 \% bins.  In contrast, human responses follow a more bell-shaped pattern, with probability mass spread across multiple bins. This supports prior findings that base LLMs lack response heterogeneity.  However, when we finetune the model on 25 \% of the first treatment group using the random sampling strategy, the distribution becomes substantially more dispersed, showing probability mass across all bins and modest peaks around 50–70 \%, closely resembling human patterns.

As shown in Figure \ref{fig:b1_b3_dist} (right panel), the human distribution for the belief that technology is deployed at both targets is highly right-skewed, with most probability mass clustered near 0–10\%. The finetuned models exhibit similar patterns as the human data and capture heterogeneity, but similar to the analysis of Target 1, the unfinetuned baseline suffers from homogeneity, with the probability mass across all simulation in the 30–50\% bins. This suggests that the unfinetuned baseline does not place proper weight on the contextual variables provided in the study (e.g., valuations, disclosure type) and instead relies on a strong, low-variance prior across the three belief options. To further investigate this, instead of analyzing only one belief at a time, we also consider the entire belief \textit{structure}, given that each participant in the experiment, and each LLM simulation, recorded these three beliefs as part of a single response. When analyzing the belief structures elicited by humans and LLM, we find that the unfinetuned LLM suffers from extreme homogeneity. As shown in Table \ref{tab:unique_q3}, the unfinetuned model produces only 19 unique belief structure  across all 929 simulations, compared to 340 in the human data. All finetuned models produce at least 200 unique combinations, indicating that even limited finetuning substantially increases the diversity of belief patterns expressed by the model.  

\begin{table}[t]
\caption{Number of unique belief structures across human data and simulation conditions.}
\centering
\begin{tabular}{l c}
\hline
\textbf{Model} & \textbf{Num. Unique Beliefs} \\
\hline
Human Data            & 340 \\
Balanced (1a1b)         & 244 \\
Random (1a1b)          & 374 \\
Balanced (All Groups)   & 338 \\
Balanced (4 Groups) & 278 \\
Random (All Groups)  & 308 \\
Random (4 Groups) & 291 \\
Unfinetuned Baseline          & 19 \\
\hline
\end{tabular}
\label{tab:unique_q3}
\end{table}

\begin{figure*}
    \centering
    \includegraphics[width=0.7\textwidth]{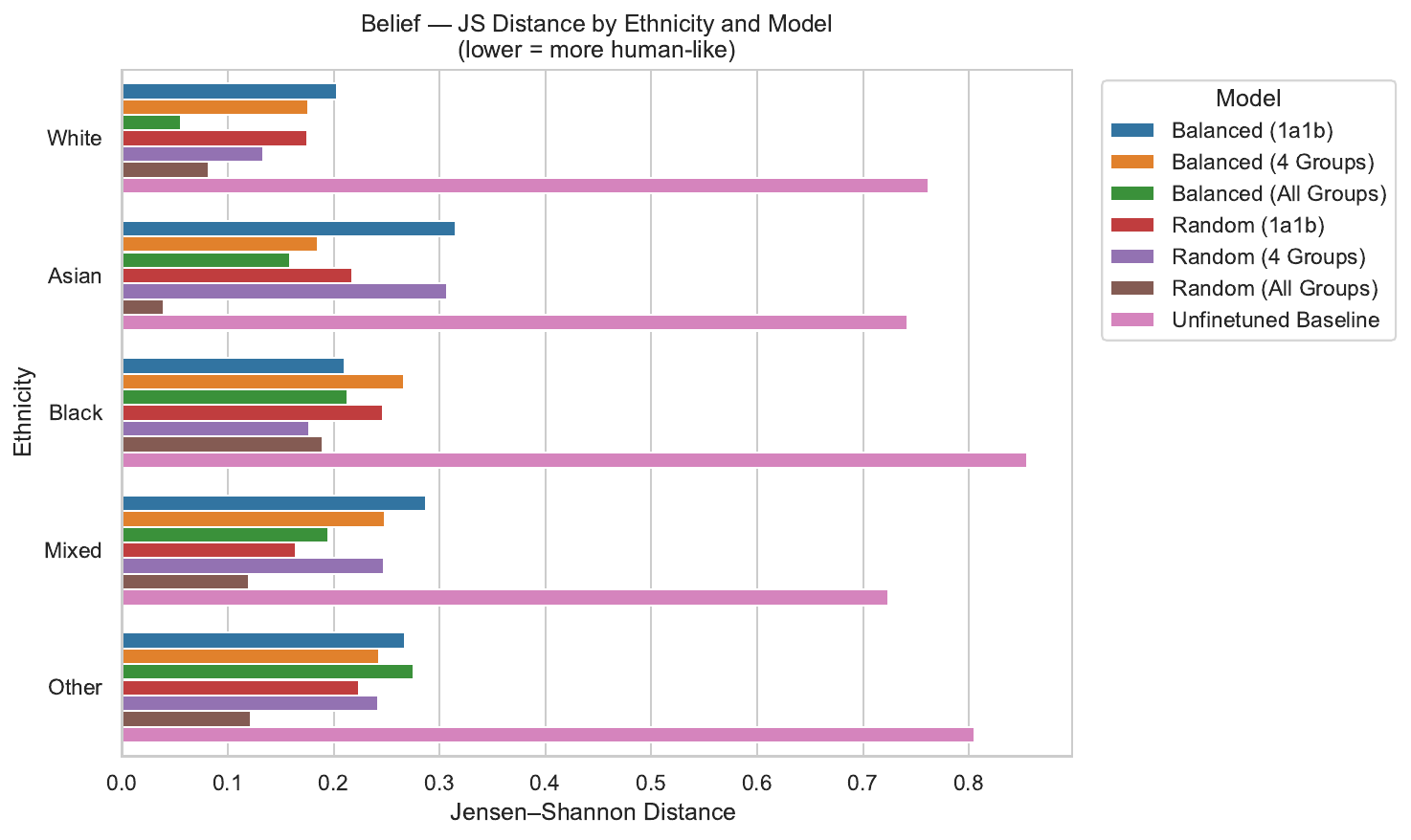}
    \caption{Jensen–Shannon distance between LLM-simulated and human belief distributions, disaggregated by ethnicity and finetuning strategy.}
    \label{fig:ethnicity_js}
\end{figure*}

\subsection{Alignment}
Given that finetuning on data subsets increases response heterogeneity, we next examine whether this added variation reflects meaningful alignment with human data or merely introduces random noise. Our results show that all finetuned models substantially reduce the Jensen–Shannon (JS) distance between LLM-simulated data and human responses. Figure \ref{fig:ethnicity_js} reports the JS distance by ethnicity and model. The ethnicity categories correspond to the default classifications used by Prolific, the participant recruitment platform through which the original data were collected. As expected, the unfinetuned baseline exhibits the largest JS distances, consistent with the patterns observed in the previous section. Across all finetuned variants, the JS distance decreases by at least half, indicating improved alignment with human response distributions. 

However, the impact of balanced sampling is mixed. Surprisingly, incorporating more demographically diverse data does not necessarily reduce JS distance for minority groups, even though such sampling increases representation. Notably, the lowest JS distance is consistently observed in the White participant category, even for the balanced model, which aligns with prior findings that LLMs most closely reflect the WEIRD population. This pattern emphasizes the persistence of demographic bias in model alignment and highlights the need for more sophisticated sampling or finetuning strategies to achieve equitable representational fidelity. Indeed, our ablation study (see Appendix \ref{sec:ablation}) shows that adding demographic information does not meaningfully increase alignment in our context.

Overall, since misalignment decreases across every subgroup, we infer that finetuning enhances overall alignment between LLM and human responses. Consistent reductions in JS distance at the group level, as presented in Appendix \ref{sec:appendix_group_js}, further corroborate this improvement. 

\begin{figure*}[!t]
    \centering
    \includegraphics[width=0.7\textwidth]{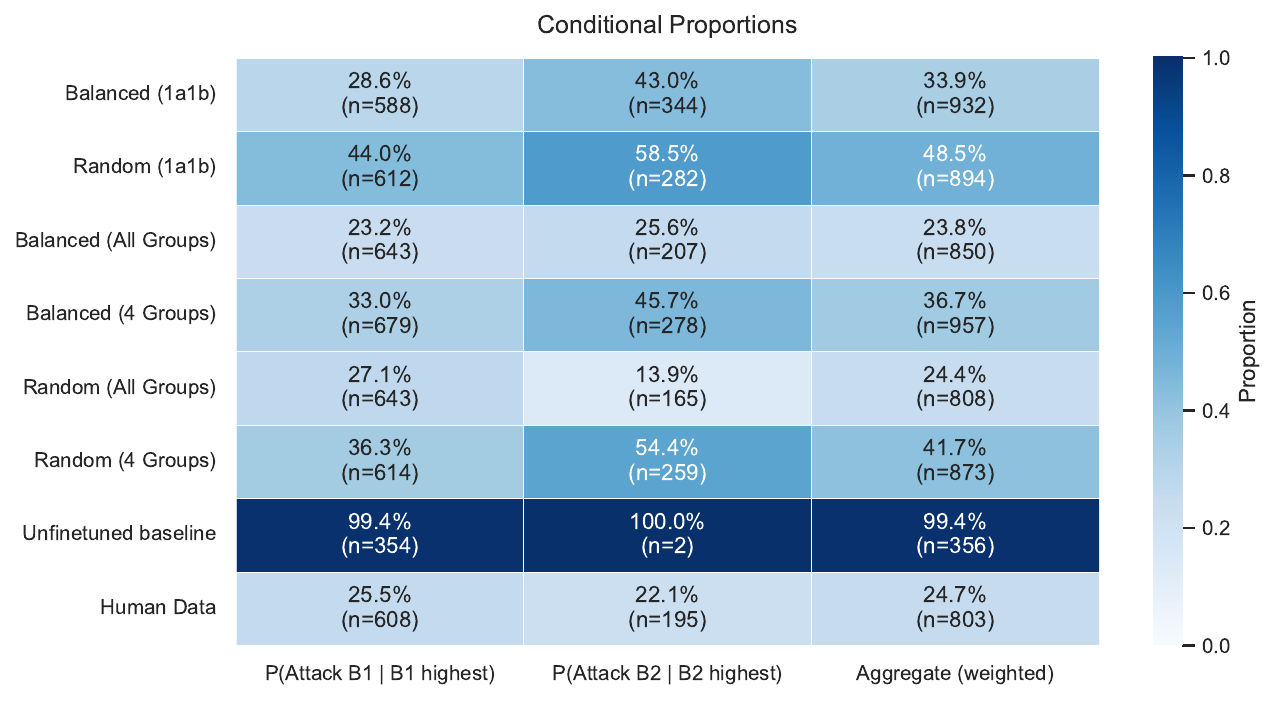}
    \caption{ Conditional proportions underlying the value–action gap. Each cell reports the proportion of cases in which the model (or humans) chooses to attack a target given that it assigns the highest belief to that target being secured, along with the corresponding sample size. Columns show: P(Attack B1 | B1 highest), P(Attack B2 | B2 highest), and the aggregate weighted proportion.}
    \label{fig:heatmap}
\end{figure*}

\subsection{Value Action Gap}
  While the original value-action gap paper \citep{shen2025mind}'s approach is theoretically grounded and conceptually novel, the authors do not benchmark their findings against the human value–action gap, which is essential for assessing how closely LLMs approximate human decision-making. In our experimental setting, we define a value–action gap as occurring when a participant, human or LLM, assigns the highest belief probability to a particular target being secured yet still chooses to attack that same target. In reality, if an attacker believes a target it highly secure, it's unlikely they would attack that target as opposed to selecting a less secure target or selecting not to attack. The unfinetuned model exhibits the largest value–action gap, almost always choosing to attack the target it assigns the highest probability of being secured. In contrast, all finetuned models display substantially smaller gaps. As shown in Figure \ref{fig:heatmap}, the value–action gap for the unfinetuned baseline is close to 100\%, whereas even models finetuned on only 25\% of a single treatment group reduce the gap by more than half. Unlike the alignment results in the previous section, where the relationship between the number of included treatment groups and JS distance was ambiguous, the value–action gap consistently shrinks as more treatment groups are incorporated into finetuning. Models trained on 25\% of all eight treatment groups attain discrepancy rates that closely match human participants (23.8\% and 24.4\% vs. 24.7\%). 

Moreover, the gap tends to be smaller for the balanced subgroup finetuning conditions than for the random sampling conditions, a pattern that did not emerge for distributional alignment. This suggests that while matching full belief distributions may require more sophisticated steering, reducing the value–action gap is comparatively easier to influence through targeted finetuning. 
\begin{table*}[t]
\caption{Comparison of coefficient recovery metrics across simulation groups.Total Coef is the total number of coefficients estimated. Reference is the number significant in the human data. Sim is the number significant in the LLM-generated data. N Match counts coefficients for which both significance and sign are correctly recovered. FPR and FNR denote false-positive and false-negative rates, and F1 (Sign) is the F1 score incorporating both significance and sign.}
\centering
\small
\begin{tabular}{lccccccc}
\toprule
\textbf{Sim Group} & \textbf{Total Coef} & \textbf{Reference} & \textbf{Sim } & \textbf{N Match} &
\textbf{FPR} & \textbf{FNR} & \textbf{F1 (Sign)} \\
\midrule
Balanced (1a1b) & 22 & 13 & 10 & 5  & 0.5555 & 0.6154 & 0.4348 \\
Random (1a1b)   & 22 & 13 & 3  & 0  & 0.1111 & 0.8462 & 0.0000 \\
Balanced (All Groups) & 22 & 13 & 15 & 9  & 0.6667 & 0.3077 & 0.6429 \\
Balanced (4 Groups) & 22 & 13 & 13 & 10 & 0.3333 & 0.2308 & 0.7692 \\
Random (All Groups)  & 22 & 13 & 12 & 7  & 0.4444 & 0.3846 & 0.6087 \\
Random (4 Groups) & 22 & 13 & 13 & 9  & 0.4444 & 0.3077 & 0.6923 \\
Unfinetuned Baseline           & 14 & 8  & 11 & 2  & 0.8333 & 0.2500 & 0.3636 \\
\bottomrule
\end{tabular}
\label{tab:coef_recovery}
\end{table*}

\subsection{Coefficient Estimation}
Given that finetuned models can increase heterogeneity, improve alignment with human data, and reduce the value–action gap, the natural next step becomes whether these simulated responses can recover the same hypothesis tests and coefficient estimates as the original human experiment. To assess this, we replicate the six regression models reported in the original study. These models fall into three conceptual categories: the first two examine whether beliefs predict an attacker’s decision to abort, the next two predict which target attackers choose, and the final two assess target choice using the normative expected value difference. 


Overall, all models produce inaccurate coefficient estimates, with the unfinetuned model and the models finetuned on only a single treatment group performing the worst. As shown in Table \ref{tab:coef_recovery}, there are 22 total coefficients across the six replicated regression models, 13 of which are statistically significant in the human data. The unfinetuned model yields only 14 coefficients because it never chooses the “abort” decision, making it impossible to estimate Models 1 and 2. 

To evaluate how well each simulated dataset recovers the human regression results, we compute F1(Sign), which jointly accounts for false positives and false negatives while also requiring that true positives have the correct sign. Across all models, the F1(Sign) values remain suboptimal. Even the best-performing finetuned model achieves an F1(Sign) of only $\approx 0.769$, and its false positive rate (FPR) remains high at $\approx 0.333$. 

These results indicate that, although finetuning on relatively small subsets of data substantially improves heterogeneity, alignment, and belief–action coherence, the resulting simulated datasets still introduce non-negligible bias when used for formal statistical inference. 

\section{Conclusion}
This study examines whether finetuning on small samples of human participant data can mitigate well-documented limitations of using base LLMs to simulate survey responses \citep{Gao, wang2025large, shen2025mind, bisbee2024synthetic}, using a behavioral experiment. Our analysis focuses on four key constructs: heterogeneity, distributional alignment, the value–action gap, and regression coefficient recovery. We find that finetuning on relatively small subsets of data substantially increases heterogeneity, improves alignment, and enhances belief–action coherence compared to the unfinetuned model, as demonstrated through both visual inspection and formal divergence metrics. Notably, heterogeneity can be substantially improved using as little as 25\% of a single treatment group, approximately 30 observations. Alignment tends to strengthen as the number of treatment groups used for finetuning increases. Unexpectedly, however, sampling equal proportions from ethnicity subgroups does not consistently enhance alignment for minority groups relative to random sampling, and misalignment remains lowest for White participants even under balanced sampling. While unexpected, this pattern aligns with prior work showing that LLMs are more closely aligned with the WEIRD population \citep{liu2022quantifying}. In contrast, when examining the value–action gap, balanced sampling performs consistently better than random sampling, suggesting that belief–action coherence may be easier to steer than subgroup-level distributional alignment, which may require more advanced finetuning strategies. 

Despite successfully mitigating several undesirable properties of the base model, our findings show that none of the finetuned models accurately replicate the regression results from the original study. This limitation indicates that full replacement of human participants remains infeasible and may introduce significant threats to the validity of empirical conclusions. Instead, we argue that LLMs finetuned on small human samples, as few as 30 observations, are better suited for in silico prototyping than base LLMs. Such models can assist researchers in refining survey instruments and conducting power or sample size analyses, but should be viewed as complements to, rather than substitutes for, human subjects \citep{anthisposition, dillion2023can, binz2025foundation}. 

\section{Discussion}
Several mechanisms may explain the misalignments between LLM-simulated and human responses. \citet{wang2025large} argue that LLMs often engage in misportraying, rather than representing the authentic behaviors of a given demographic group, models reproduce what out-group members believe those in-group members would do. Since demographic identity is rarely encoded directly in the text on which LLMs are trained, explicit mentions of identity may instead express stereotypes by out-group authors. \citet{liu2024large} offer a related explanation, and it aligns with the sociolingustic theoretical model introduced by \citet{zhou2025culture}. They study LLM performance on datasets involving risky decision-making and find that LLMs align more closely with people’s expectations of how others behave than with how individuals actually behave. Since people tend to assume that others are more rational than themselves, LLMs may encode these perceptions rather than genuine human behavior. These findings suggest that LLM decisions may be shaped less by representations of actual in-group behavior and more by culturally shared beliefs about what “typical” behavior should look like. Additional factors may also contribute to misalignment. The training data themselves may contain survey responses affected by social desirability bias \citep{crowne1960new}, meaning that models learn socially acceptable patterns of responding rather than authentic responses. More directly, reinforcement learning from human feedback (RLHF) \citep{ouyang2022training} optimizes models to behave in ways that humans prefer LLMs to behave \citep{dahlgren2025helpful}, not necessarily in ways that mirror empirical human data. 

A concurrent work by \citet{krsteski2025valid} also examines finetuning on a small human survey sample, but in a different setting. They study panel surveys and use earlier waves as historical data to finetune or prompt LLMs and generate synthetic responses for a later target wave. A small subsample of human responses from that target wave is then used for statistical rectification of the model-based estimates, and their main focus is how to optimally allocate limited human responses between finetuning and post-hoc correction to reduce bias in LLM-based survey simulations. By contrast, our study does not focus on bias correction, rather, we evaluate whether finetuning on small samples can recover key behavioral constructs within a controlled experimental setting. 

\section{Limitations}
This study has several limitations. First, we rely on a single case study from one behavioral domain, and our findings may not generalize to other experimental settings or survey contexts. Second, since finetuning was performed multiple times, all comparisons were conducted using a single base model. Future work could assess whether our results hold across different model families or architectures. Third, due to imbalance in the original dataset, our balanced finetuning sets could only include up to 25\% of the data without serious violation of the equal-proportion constraint, which may limit the sample points available for minority subgroups. Finally, to remain faithful to the original experimental design, we did not systematically explore the sensitivity of LLMs to prompt phrasing, format, or ordering \citep{tjuatja2024llms, dominguez2024questioning}, factors that could meaningfully influence model behavior and should be examined in future research. 
\section{Ethics Statement}
Although this study explores the simulation of human responses using LLMs, we do not advocate replacing human participants with LLMs for informing policy, healthcare, or other high-stakes decisions, particularly those affecting disadvantaged or marginalized populations, given the limitations identified in this work and in numerous prior studies. We adhere to the ACL Code of Ethics throughout the research process and comply with OpenAI’s usage policies when using GPT models and finetuning tools. We ensured that the data from the original experiment contain no harmful or stereotypical content. We also confirmed with the original author that the study received IRB approval from their institution. All datasets and code will be released for academic use in accordance with ethical research guidelines.

\bibliography{custom}
\clearpage
\appendix

\section{Prompt}
\label{sec:prompt}

\subsection{Prompt 1}
\{Background Info\} Based on your preliminary surveillance, you have identified the following information: The value you place on the treasure at Beach 1: \{Your Valuation 1\} coins
        The value you place on the treasure at Beach 2: \{Your Valuation 1\} coins
        The value Queen Katrina places on the treasure at Beach 1: \{Queen's Valuation 1\} coins
        The value Queen Katrina places on the treasure at Beach 2: \{Queen's Valuation 2\} coins
        Therefore, if you successfully rob either beach, you would have \{Reward Value\} coins worth of treasure. There will a fixed cost for you to rob the beaches given that you will need 
        to travel to Gold Island and execute the robbery. These costs are as follows:
        Cost to rob Beach 1: \{Cost 1\} coins
        Cost to rob Beach 2: \{Cost 2\} coins
        As you were watching the news today, you learned that Queen Katrina would be making a big announcement later this afternoon. You have decided to wait until you hear what she 
        has to say before planning and executing your big robbery.
        Based on this initial information, which beach would you choose to rob? Output your choice (beach 1 or beach 2) inside square brackets (array 1)
        Final output 
        should only contain array one.
\subsection{Prompt 2}
\label{subsec:prompt2}
At this point, you will read Queen Katrina’s announcement. Her announcement is in regards to new security measures that have been deployed to protect the treasure.
You know that the new security measures could be deployed only at Beach 1, only at Beach 2, or at both beaches. If the new security measures are actually deployed at Beach 1, then the chance of a robbery being successful there is 30\%. Similarly, if the new security measures are actually deployed at Beach 2, then the chance of a robbery being successful there is also 30\%. The problem is that only Queen Katrina knows exactly where the new security measures are deployed. Furthermore, the information she releases could be totally true, totally false, or partially false (i.e., some parts of her announcement could be true, while other parts could be false).
After reading her announcement, you will be asked a series of questions regarding your beliefs about the security measures, and your plans for the robbery.
Queen Katrina’s announcement:
\{Queen's Announcement\}
treasure.
Reminder (data from previous page):
The value Queen Katrina places on the treasure at Beach 1: \{Queen's Valuation 1\} coins
The value Queen Katrina places on the treasure at Beach 2: \{Queen's Valuation 2\} coins
What is the chance (in your belief) that the new security is deployed only at Beach 1, only at Beach 2, or both? Choose a value between 0–100\% (note: the sum of all 3 prompts must equal 100\%):
Output your beliefs (3 percentages that add up to 100\%) inside square brackets (array 2)

What influenced your responses to the question? Choose all options that apply:
Option 1: The information that Queen Katrina released in her announcement
Option 2: Queen Katrina’s values of the treasure at the different beaches
Output your answer inside square brackets (array 3)

Aside from the base pay for this study, you are now receiving more money which you can use to fund your robbery. In specific, 1 coin in the game is equivalent to 0.000001 USD (United States dollars). For example, it costs 150,000 coins to rob Beach 2 in this game, which is equivalent to \$0.15. You now have \$0.15 that you can use to fund your robbery.
If you choose not to rob either beach, you keep the \$0.15.
If you choose to rob Beach 1, you will spend \$0.10 of the \$0.15, and you keep the remaining \$0.05. If you succeed in robbing Beach 1, you will receive \$0.80 in additional money. If you fail, you will receive \$0.00 in additional money.
If you choose to rob Beach 2, you will spend the total \$0.15. If you succeed in robbing Beach 2, you will receive \$0.80 in additional money. If you fail, you will receive \$0.00 in additional money.
Reminders
The probability of successfully robbing a beach if the new security measures are deployed is 30\%, and the probability of successfully robbing a beach if the new security measures are not deployed is 80
The value you place on the treasure at Beach 1: \{Your Valuation 1\} coins
The value you place on the treasure at Beach 2: \{Your Valuation 2\} coins
Cost to rob Beach 1: \{Cost 1\} coins
Cost to rob Beach 2: \{Cost 2\} coins
The value Queen Katrina places on the treasure at Beach 1: \{Queen's Valuation 1\} coins
The value Queen Katrina places on the treasure at Beach 2: \{Queen's Valuation 2\} coins
What will be your final plan for the robbery? Choose one of the following responses:
Option 1: Attempt to steal the treasure from Beach 1
Option 2: Attempt to steal the treasure from Beach 2
Option 3: Abort the heist altogether and do not attempt to steal treasure from either beach
Output your decision (only 1, 2, or 3) inside square brackets (array 4)
Briefly explain your response to the previous question (50 words maximum)
Output your explanation inside square brackets (array 5)
Final output should only contain array 2, 3, 4, 5, separated by a comma.

\section{Experimental Manipulations}\label{sec:manipulations}
\small
\begin{table}[h!]
\caption{Manipulations from the experiments in \citet{Hunt}}
\label{tab:treatments}
\centering
\small
\begin{tabular}{|p{1.1cm}|p{6cm}|}
\hline
Notation & Explanation\\	
\hline
V1 & Large difference in the defender's target valuations (\$1.2 million vs. \$400,000)\\
\hline
V2 & Small difference in the defender's target valuations (\$850,000 vs. \$750,000) \\
\hhline{|=|=|}
A1 & Information released regarding the deployment of the new defense measures; \textbf{no target-level information released} \\
\hline
A2 & Information released regarding the deployment of the new defense measures \textit{and} that the defense measures are deployed at \textbf{Target 1} \\
\hline
A3 & Information released regarding the deployment of the new defense measures \textit{and} that the defense measures are deployed at \textbf{Target 2} \\
\hline
A4 & Information released regarding the deployment of the new defense measures \textit{and} that the defense measures are deployed at \textbf{Targets 1 and 2}\\
\hline
\end{tabular}
\end{table}

\section{Belief}
\label{sec:appendix_belief}
\begin{figure*}
    \centering
    \includegraphics[width=0.48\textwidth]{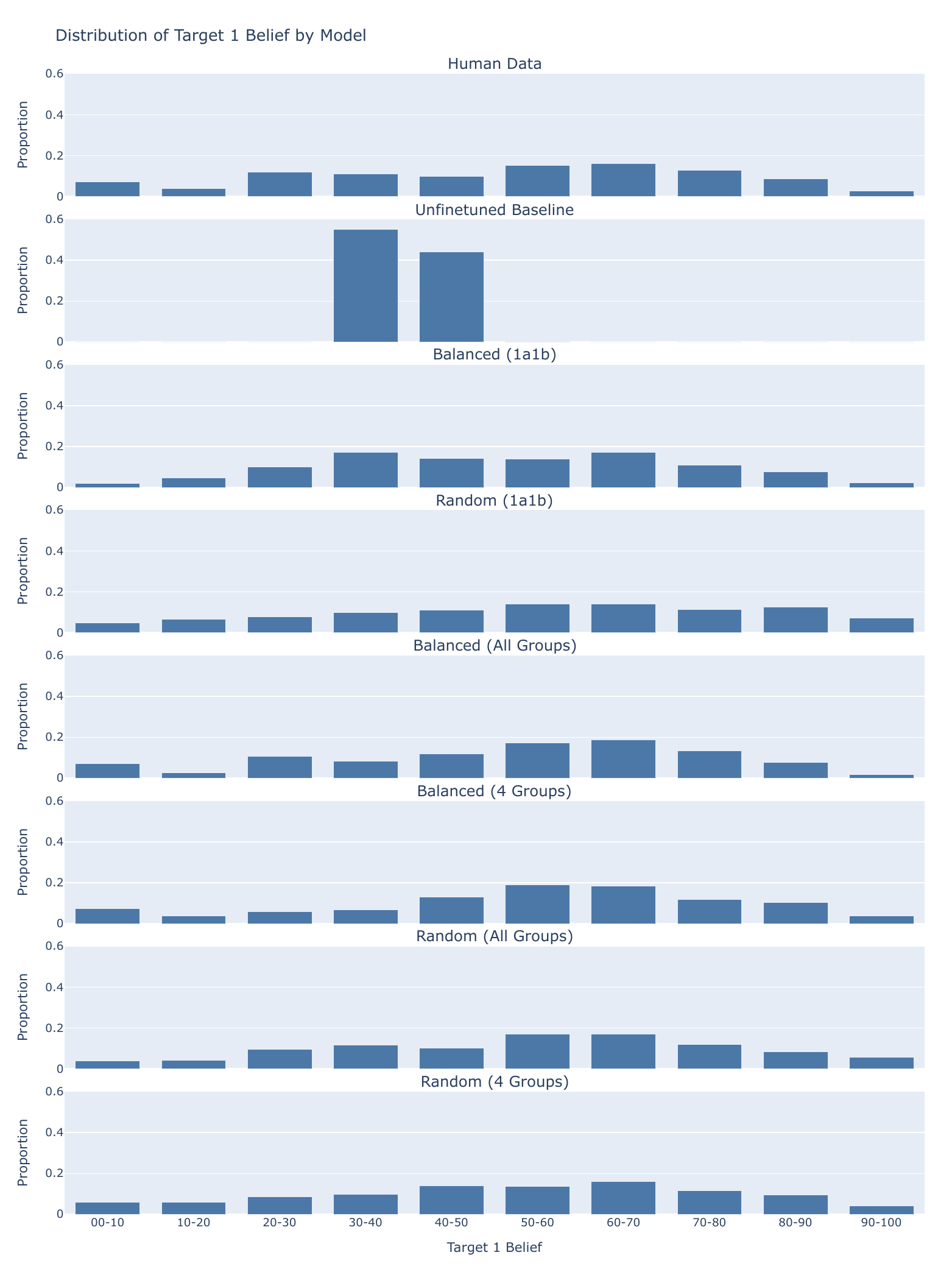}\hfill
    \includegraphics[width=0.48\textwidth]{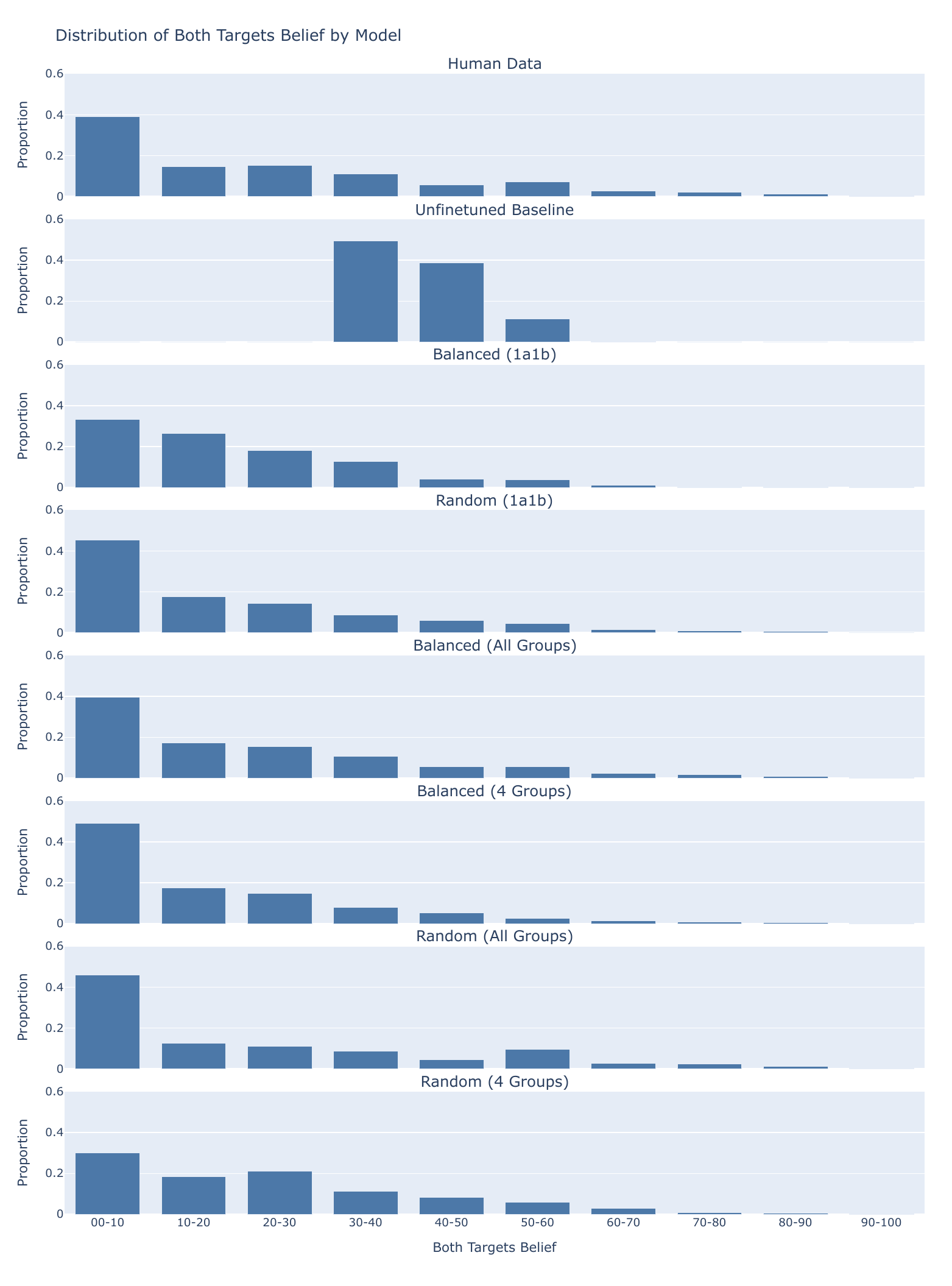}
    \caption{Distributions of beliefs that security is deployed at Target 1 (left) and at both targets (right) across human data and LLM simulation conditions. Each panel shows the empirical belief distribution for human participants alongside models finetuned under different sampling strategies.}
    \label{fig:b1_b3_dist_full}
\end{figure*}
As shown in Figure \ref{fig:b1_b3_dist_full}, expanding finetuning to include randomly sampled data from four treatment groups ("Random (4 Groups)") and all eight groups ("Random (All Groups)") further improves diversity. The balanced variants, "Balanced (1a1b)," "Balanced (4 Groups)," and "Balanced (All Groups)," also show similarly enhanced diversity and alignment. 

\begin{figure*}[t]
  \includegraphics[width=0.7\textwidth]{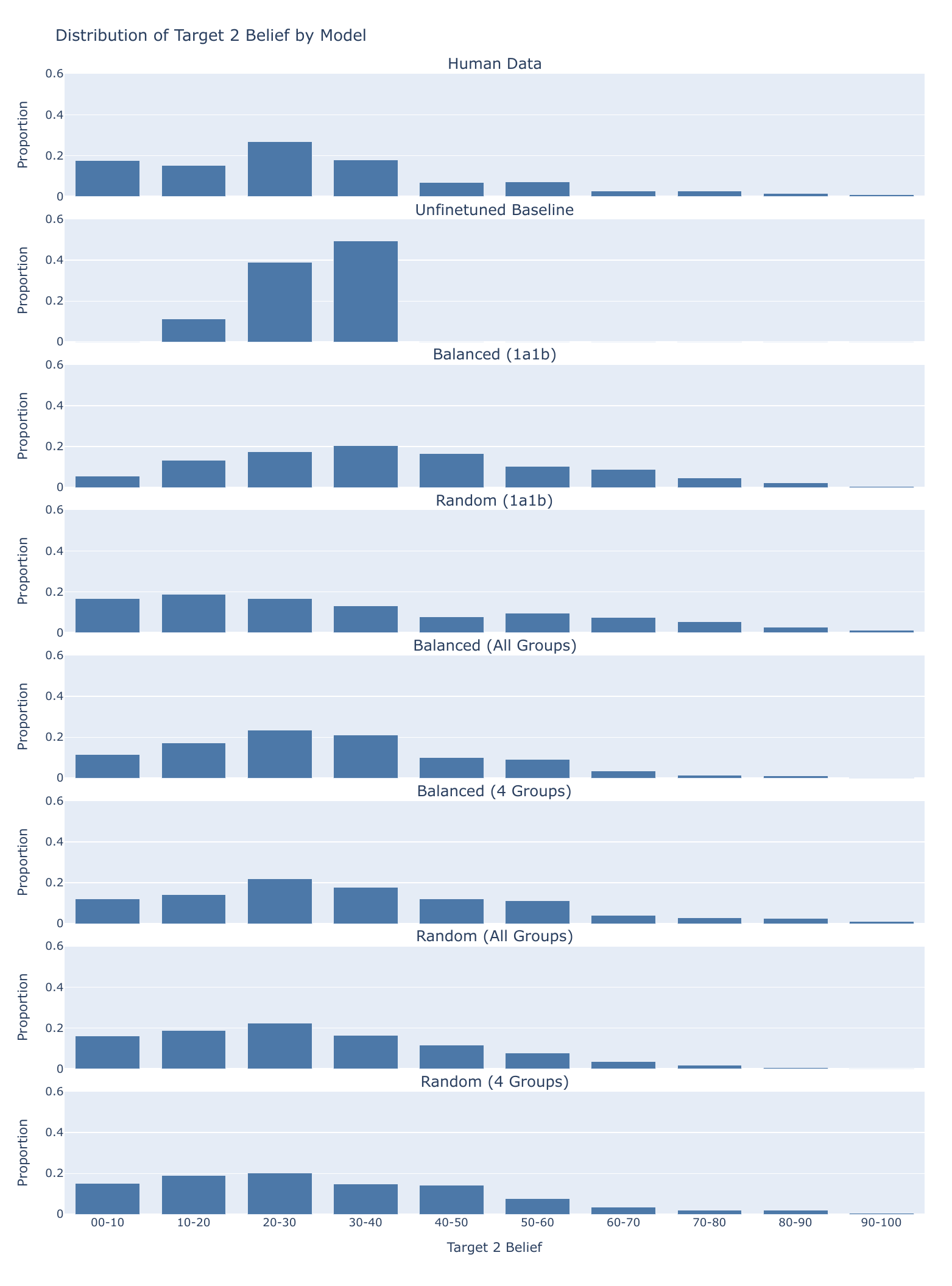} 
  
  \caption {Distribution of beliefs for Targets 2 across human data and LLM simulation conditions. Each panel shows the empirical belief distribution for human participants alongside models finetuned under different sampling strategies.}
  \label{b2_dist}
\end{figure*}

The distributions of the finetuned models for Targets 2 closely resemble the human data. As shown in Figure \ref{b2_dist}, they are right-skewed, with most probability mass concentrated in the lower ranges of the distribution. In contrast, the unfinetuned baseline shows a highly concentrated pattern, placing nearly all probability mass in the 20–40\% range and assigning no probability beyond 40\%. Compared to the belief distributions for Targets 1 and Both Targets, the peak of the unfinetuned model for Targets 2 is actually closer to the human data. Both the human data and the unfinetuned model place their highest probability mass in the 30–40\% range. The human distribution is bimodal, with a secondary peak in the 0–10\% bin. In contrast, most of the finetuned models exhibit a Gaussian-like, unimodal distribution centered around the middle bins. The exception is Random (1a1b), which remains noticeably right-skewed.

\section{Group-Level JS}
\label{sec:appendix_group_js}
\begin{figure*}
  \includegraphics[width=0.7\textwidth]{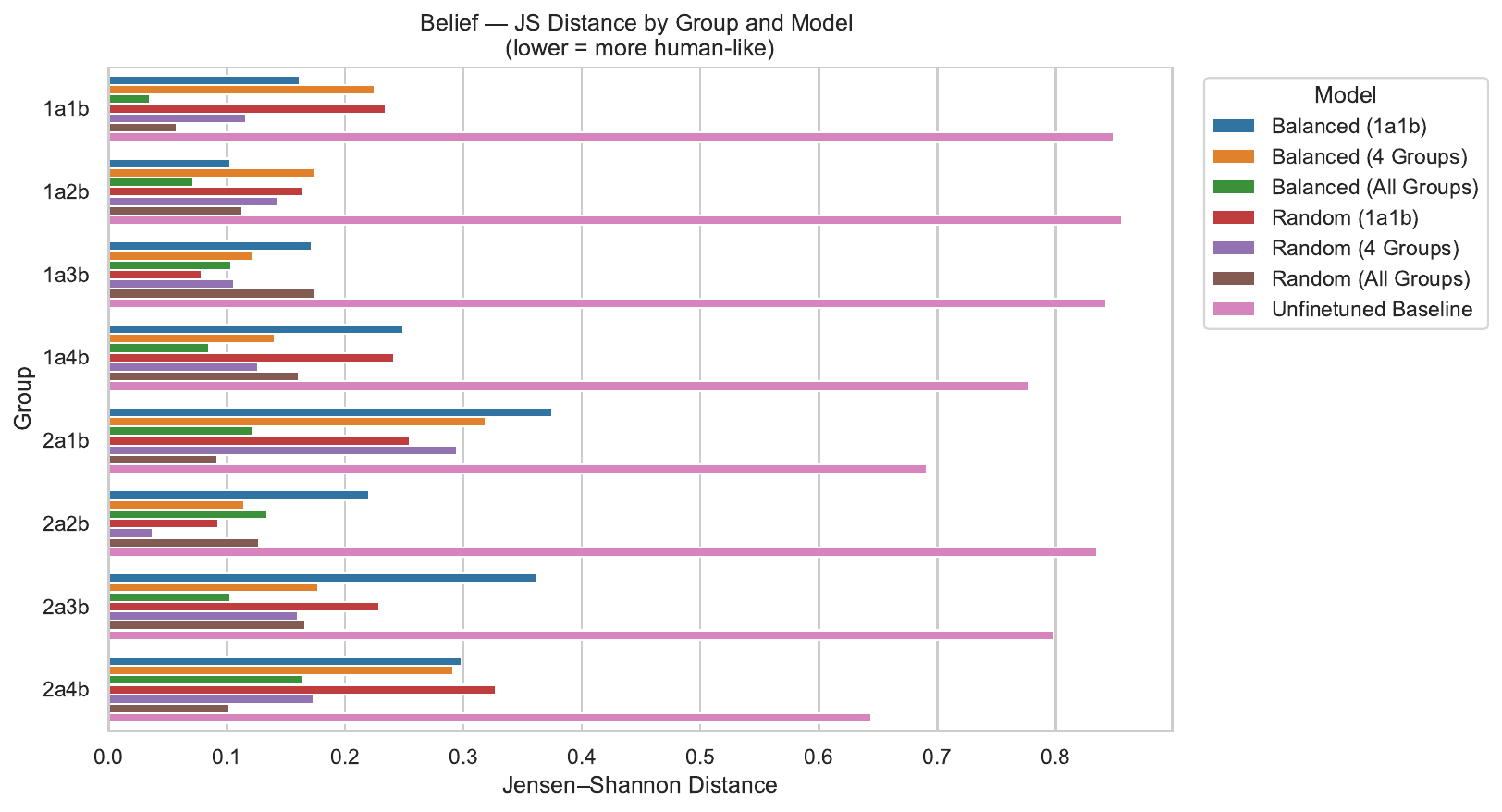} 
  
  \caption {Jensen–Shannon (JS) distance between LLM-simulated and human belief distributions by treatment group and finetuning strategy.}
  \label{group_js}
\end{figure*}

At the group level, the unfinetuned model continues to show the largest divergence from the human data, whereas all finetuned models achieve at least roughly half the JS distance of the baseline. As shown in Figure \ref{group_js}, the reduction in JS distance is mildly correlated with the number of treatment groups included in the finetuning data: models finetuned on all eight groups perform slightly better than those trained on four groups, which in turn outperform the one-group models. The models finetuned on the first four groups exhibit somewhat lower divergence within those same groups—an expected pattern given their training data, but the differences are small. This suggests that the finetuned models retain a reasonable degree of generalizability beyond the specific treatment groups they were trained on.

\section {Ablation JS}
\label{sec:ablation}
\begin{figure*}
  \includegraphics[width=0.7\textwidth]{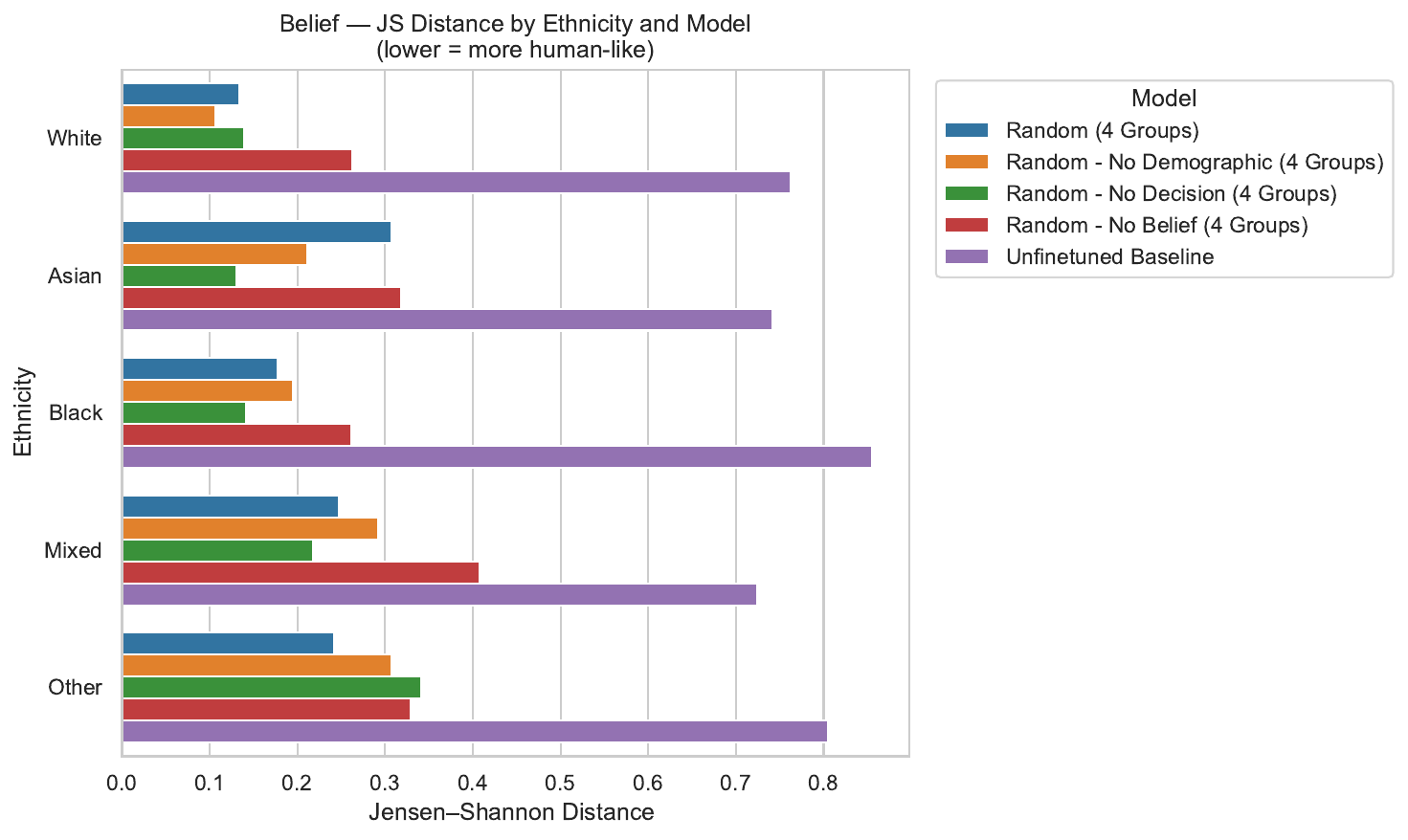} 
  
  \caption{ Ablation study of a finetuned LLM showing Jensen–Shannon (JS) distances between simulated and human belief distributions by ethnicity (lower values indicate closer alignment). “Random – No Demographic,” “Random – No Decision,” and “Random – No Belief” exclude demographic background, decision responses, and belief responses, respectively, during finetuning. The unfinetuned baseline is shown for comparison.}
  \label{group_js_ablation}
\end{figure*}

Figure \ref{group_js_ablation} presents an ablation study conducted on one of our finetuned models. Random – No Demographic refers to a variant trained without participants’ background information. Random – No Decision and Random – No Belief refer to variants that exclude participants’ reported final decisions and belief responses, respectively, during finetuning. Due to budget constraints, we perform this ablation on a single representative model (Random (4 Groups)) rather than across all finetuned models in the manuscript.  As shown in Figure \ref{group_js_ablation}, the Jensen–Shannon (JS) distances for Random – No Demographic are similar to those of the original finetuned model across ethnic subgroups, suggesting that including demographic background information alone does not meaningfully reduce subgroup misalignment in belief distributions. In contrast, Random – No Belief exhibits larger JS distances than the other finetuned variants, indicating increased misalignment when decision information is excluded, nevertheless, its performance remains substantially better than that of the unfinetuned baseline. Finally, Random – No Decision performs comparably to the original model, implying that adding decision information does not indirectly improve belief alignment when belief data themselves are omitted during finetuning. 

\end{document}